\documentclass[a4paper]{article}
\usepackage{dx}
\usepackage[T1]{fontenc}
\usepackage{ae,aecompl}
\usepackage{amsfonts}
\usepackage{amsmath}
\usepackage{amsthm}

\usepackage{times}

\usepackage[inline]{enumitem}

\usepackage{rotating}

\usepackage[table]{xcolor}    
\usepackage{scalerel}
\usepackage{array,multirow}

\newcommand{\OK}{\textsc{ok}}
\newcommand{\NOK}{\textsc{nok}}
\newcommand{\md}{\mathcal{D}}

\newcommand{\mc}{\mathcal{C}}

\newcommand{\sd}{\textsc{sd}}
\newcommand{\comps}{\textsc{comps}}
\newcommand{\obs}{\textsc{obs}}
\newcommand{\meas}{\textsc{meas}}
\newcommand{\ok}{\textsc{ok}}

\newtheorem{prob}{Problem}

\title{How should I compute my candidates? \\
A taxonomy and classification of diagnosis computation algorithms\thanks{Supported by Austrian Science Fund, contract P-32445-N38.}\thanks{This work represents an earlier version of \protect\cite{rodler_ecai2023} which appeared in the proceedings of the \emph{European Conference on Artificial Intelligence (ECAI) 2023}. Please refer to the ECAI-version.}}
\author%
{%
Patrick Rodler 
\\
University of Klagenfurt\\
e-mail: patrick.rodler@aau.at
}

\begin{document}

\maketitle

\begin{abstract}
  This work proposes a taxonomy for diagnosis computation methods which allows their standardized assessment, classification and comparison.
  The aim is to 
  \emph{(i)}~give researchers and practitioners an impression of the diverse landscape of available diagnostic techniques, 
  \emph{(ii)}~allow them to easily retrieve the main features as well as pros and cons of the approaches, 
  \emph{(iii)}~enable an easy and clear comparison of the techniques based on their characteristics wrt.\ a list of important and well-defined properties, and
  \emph{(iv)}~facilitate the selection of the ``right'' algorithm to adopt for a particular problem case, e.g., in practical diagnostic settings, 
  for comparison in experimental evaluations, or for reuse, modification, extension, or improvement in the course of research. 
\end{abstract}


\section{Introduction}
Diagnosis computation is one of the most integral tasks in model-based diagnosis as it allows to generate fault hypotheses, which are essential for both fault localization and repair. 
Due to its generality, the model-based diagnosis formalism
has been used to express and tackle a wide diversity of 
debugging problems in application areas ranging from software \cite{hunt_model-based_1998}, logic programming \cite{eiter_data_nodate}, recommender systems \cite{felfernig_automated_2007}, ontologies \cite{shchekotykhin_interactive_2012} and knowledge bases \cite{rodler_rio_2013} via hardware \cite{friedrich_model-based_1999}, circuits \cite{de_kleer_diagnosing_1987} and robots \cite{zaman_integrated_2013} to scheduling \cite{rodler_randomized_2021}, aircrafts \cite{gorinevsky_model-based_2002} and cars \cite{sachenbacher_electrics_1998}. 
This has led to a remarkable multitude and heterogeneity of the diagnosis computation methods proposed in literature, which are often motivated by and tailored for application-specific requirements and problem cases.
As a result, it is a hard task for researchers and practitioners alike to
\begin{itemize}[noitemsep,topsep=0pt]
	\item get an overview of existing approaches,
	\item identify the crucial properties of diagnostic techniques, 
	\item assess
	the methods based on these properties, 
	and 
	\item choose the appropriate approach for a (research- or application-related) diagnostic task at hand.
\end{itemize} 
With this work, we account for this by presenting a taxonomy for diagnosis computation algorithms. Specifically, we introduce and formally define a range of features which are arguably vital for a proper understanding, comparison, selection, and use of diagnostic techniques. We explain the influence of each feature on the proper selection of a diagnosis algorithm for a diagnostic task, discuss the potential impact of different feature manifestations on the performance of diagnosis algorithms, and examine relationships among the features. To demonstrate the value and application of the proposed taxonomy, we provide a multi-dimensional assessment and categorization of several important diagnostic methods in the literature.

\section{Preliminaries}
\label{sec:basics}
\emph{Model-based diagnosis} \cite{reiter_theory_1987}
assumes a system (e.g., software, circuit, knowledge base, physical device) consisting of a set of \emph{components} $\comps=\{c_1,\dots,c_n\}$ (e.g., lines of code, gates, axioms, physical constituents) which is formally described 
in some monotonic logical language. Beside any relevant general knowledge about the system, this \emph{system description} $\textsc{sd}$ includes a specification of the normal behavior (logical sentence $\textsc{beh}(c_i)$) of all 
components $c_i \in \comps$ of the form $\textsc{ok}(c_i) \to \textsc{beh}(c_i)$. As a result, when assuming all components to be fault-free, i.e., $\OK(\comps) := \{\ok(c_1), \dots, \ok(c_n)\}$, conclusions about the normal system behavior 
can be drawn by means of 
theorem provers. When the real system behavior, ascertained through \emph{system observations} 
and/or \emph{system measurements} (stated as logical sentences $\obs$ and $\meas$),
is inconsistent with the system behavior predicted by $\sd$, the normality-assumption for some of the components has to be retracted. We refer to $\langle\sd,\comps,\obs,\meas\rangle$ as a \emph{diagnosis problem instance (DPI)}. 

For a DPI, one is interested in finding a diagnosis, i.e., a set of components whose abnormality would explain the observed incorrect system behavior. Formally, a set of components $\md \subseteq \comps$ is called a \emph{diagnosis} iff $\sd \cup \obs \cup \meas \cup \OK(\comps\setminus\md) \cup \NOK(\md)$ is consistent where $\OK(X) := \{\ok(c_i) \,|\, c_i \in X\}$ and $\NOK(X) := \{\lnot \ok(c_i) \,|\, c_i \in X\}$. A diagnosis $\md$ is termed \emph{minimal} / \emph{minimum-cardinality} iff there is no diagnosis $\md'$ such that $\md'\subset\md$ / $|\md'|<|\md|$. For efficiency and tractability reasons, the focus in model-based diagnosis is often laid on minimal diagnoses only \cite{kleer_focusing_1991}. In particular, the minimal diagnoses are representative of all diagnoses under the \emph{weak fault model} \cite{de_kleer_characterizing_1992}, where the system description $\sd$ contains only information about the normal behavior of the system components (and leaves the components' behavior undefined in case of failure). We restrict the study in this paper to diagnosis computation methods relying on a weak fault model. They address the following problem:
\begin{prob}[Diagnosis Computation]\label{prob:diag_computation}\phantom{.}\\
	\textbf{Given:} A DPI $\langle\sd,\comps,\obs,\meas\rangle$, an integer $k \geq 1$. \\
	\textbf{Find:} Find $k$ minimal diagnoses (satisfying a property $p$) for $\langle\sd,\comps,\obs,\meas\rangle$.
\end{prob}
Diagnostic techniques may solve different manifestations of this problem. E.g., they might aim at computing $n$ / all minimal diagnoses (by specifying $k:=n$ / $k := \infty$), or at finding all minimum-cardinality diagnoses (by specifying $k := \infty$ and $p:=$ minimum-cardinality).

Useful for diagnosis computation and technically closely related to the concept of a diagnosis is the notion of a conflict, which is a set of components such that the assumption of all of them being fault-free is inconsistent with the current knowledge about the system. Formally, a set of components $\mc \subseteq \comps$ is a \emph{conflict} iff $\sd \cup \obs \cup \meas \cup \OK(\mc)$ is inconsistent.
Again, we call a conflict $\mc$ \emph{minimal} iff there is no conflict $\mc'$ with $\mc' \subset \mc$. 
There are two important links between diagnoses and conflicts \cite{reiter_theory_1987}: 
\begin{description}[font=\normalfont\em,noitemsep,topsep=0pt]
	\item[Hitting-set Property] A (minimal) diagnosis is a (minimal) hitting set of all minimal conflicts. \\
	($X$ is a \emph{hitting set} of a collection of sets $\mathbf{S}$ iff $X \subseteq \bigcup_{S \in \mathbf{S}} S$ and $X \cap S \neq \emptyset$ for all $S \in \mathbf{S}$.)
	\item[Duality Property] $X$ is a diagnosis 
	iff $\comps\setminus X$ is not a conflict. 
\end{description}
  
In many cases, there is a substantial number of competing diagnoses for a given DPI. The goal is then to isolate the \emph{actual diagnosis} which pinpoints the \emph{actually} faulty components. Basically, two strategies exist to handle multiple diagnoses, aiming at reducing or avoiding the effort for a manual inspection of the diagnoses: \emph{(i)}~\emph{rank or restrict the computed diagnoses} based on some informative criterion such as maximal probability or minimal cardinality \cite{kleer_focusing_1991}, or \emph{(ii)}~\emph{apply sequential diagnosis techniques} to acquire additional information about the diagnosed system to gradually refine the set of diagnoses \cite{de_kleer_diagnosing_1987}. 
Whereas rankings or focusing techniques can be very powerful if the actual diagnosis appears (early) in the solution list,
there is no guarantee that the actually faulty components will be located (efficiently). 

The more sophisticated sequential diagnosis techniques, on the other hand, 
gather further system measurements ($\meas$) to iteratively rule out spurious diagnoses. They aim at solving the following problem (with highest efficiency): 
\begin{prob}[Sequential Diagnosis]\label{prob:seq_diagnosis} \phantom{.}\\
	\textbf{Given:} A DPI $\langle\sd,\comps,\obs,\meas\rangle$. \\
	\textbf{Find:}  A sequence (set) of measurements, expressed as logical sentences $m_1,\dots,m_k$, such that there is a single (highly probable) minimal diagnosis for $\langle\sd,\comps,\obs,\meas \cup \{m_1,\dots,m_k\}\rangle$.
\end{prob}
Many sequential diagnosis methods can be characterized by a recurring execution of four phases \cite{de_kleer_how_1993}: 
\emph{(1)}~computation of a set of (preferred, e.g.,
most probable) minimal diagnoses, \emph{(2)}~selection of the most informative system measurement based on these, \emph{(3)}~conduction of measurement actions (by some user or oracle, e.g., an electrical engineer if a circuit is diagnosed), and \emph{(4)}~exploitation of the measurement outcome to update 
the system knowledge. That is, the DPI is modified (by extending $\meas$) between each two 
iterations of these phases. The execution stops 
if Problem~\ref{prob:seq_diagnosis} is solved, i.e., sufficient diagnostic certainty 
is achieved. 

\section{A Taxonomy for Diagnosis Algorithms}
\label{sec:features}
In this section, we propose a collection of pivotal features of diagnosis computation algorithms, based on which we will classify and compare some important existing techniques in Sec.~\ref{sec:classification} and Tab.~\ref{tab:algo_classification}. 
In the following, we assume that an algorithm $A$ addresses (some manifestation of) the diagnosis computation problem (cf.\ Problem~\ref{prob:diag_computation}) and is given as input a DPI and possibly some meta information (such as component failure rates that allow to derive diagnosis probabilities \cite{de_kleer_diagnosing_1987}, or algorithm-specific parameters, e.g., stop, pruning or restart criteria \cite{feldman_computing_2008,abreu_low-cost_2009}).\footnote{Works describing diagnosis algorithms use a wide variety of notations and formalisms, which can however also be expressed by means of Reiter's general theory \cite{reiter_theory_1987}, as reviewed in Sec.~\ref{sec:basics}.} We describe each feature by giving a \emph{definition} of its possible manifestations, a brief explanation of its \emph{relevance} wrt.\ algorithm selection for a diagnostic task, a short discussion of the practical \emph{impact} of different feature manifestations, and a comment on the \emph{relationship} to other features. 
The features can be logically grouped into five categories, i.e., 
\emph{Output Qualities} (Bullets~\ref{enum:features:soundness}--\ref{enum:features:type_of_output}), \emph{Way of Computation} (\ref{enum:features:conflict-based}--\ref{enum:features:way_of_conflict_computation}), \emph{Sequential Diagnosis Context} (\ref{enum:features:focus_on_sequential_diagnosis}--\ref{enum:features:maintenance_of_state}), \emph{Application Context} (\ref{enum:features:general_applicability}--\ref{enum:features:logics-agnosticism}), and \emph{Performance} (\ref{enum:features:space_efficiency}), as shown in Tab.~\ref{tab:algo_classification}:
\begin{enumerate}[noitemsep,topsep=0pt]  
	\item \label{enum:features:soundness}\textbf{Soundness:} \\
	\underline{\emph{Definition:}} $A$ is \emph{sound} iff it outputs only minimal diagnoses; otherwise, it is \emph{unsound}.\\
	\underline{\emph{Relevance:}} Soundness is necessary if \emph{(a)}~no actually fault-free components should be marked as faulty in a diagnostic scenario, e.g., when inspecting or changing parts unnecessarily is costly such as in a car \cite{heckerman_decision-theoretic_1995}, or when a modification of correct components impacts the quality of the system such as for axioms in a knowledge base \cite{rodler_interactive_2015}, or if \emph{(b)}~every solution returned by the algorithm should indeed be a possible explanation for the observed system misbehavior, e.g., to avoid the necessity of additional computations and of a potentially costly post-processing of the solutions. Apart from that, the soundness requirement is in line with the generally accepted principles of Parsimony \cite{reiter_theory_1987} and Occam's razor \cite{blumer_occams_1987}, which postulate that from two different (fault) explanations, the simpler one is preferable.\\ 
	\underline{\emph{Impact:}} Forgoing the requirement of soundness can lead to a higher efficiency of diagnosis computation, as certain unsound algorithms are designed to drop soundness to the benefit of performance (e.g., \cite{li_computing_2002,feldman_computing_2008}). There are basically two forms of unsoundness for returned diagnoses, i.e., they might be \emph{(a)}~non-minimal diagnoses (intuitively: ``too large'' component sets; cf., e.g., \cite{feldman_computing_2008}), or \emph{(b)}~non-diagnoses (intuitively: ``too small'' component sets; cf., e.g., \cite{li_computing_2002}). Both cases can be handled by a suitable post-processing of the returned solutions (cf., e.g., \cite{jiang_computing_2002}), 
	the cost of which depends on the number of solutions that are non-(minimal) diagnoses and on their degree of unsoundness (i.e., how much ``too small'' or ``too large'' the diagnoses are).\\
	\underline{\emph{Relationship:}} Unsoundness can entail incompleteness or a violation of the best-first property, e.g., for systematic hitting set searches (e.g., \cite{rodler_interactive_2015}).  
	\item \label{enum:features:completeness}\textbf{Completeness:} \\
	\underline{\emph{Definition:}} If $A$ computes a set of diagnoses, it is \emph{all-complete} iff it outputs all minimal diagnoses given sufficient time and memory, and it is \emph{property-complete} iff it outputs all minimal diagnoses with a particular property (e.g., minimum cardinality) given sufficient time and memory. If $A$ computes a single diagnosis (with a particular property $p$, e.g., minimum-cardinality), it is \emph{one-complete} iff it outputs a minimal diagnosis (with property $p$) whenever such a diagnosis exists. 
	Otherwise, if there is any minimal diagnosis that $A$ might fail to compute, 
	it is \emph{incomplete}.\\
	\underline{\emph{Relevance:}} Completeness is necessary if it is crucial in a diagnostic scenario that the actual diagnosis is found with certainty, or when missing the actual diagnosis or a diagnosis with a particular property might have serious consequences, e.g., when diagnosing critical systems such as aircrafts, medical ontologies or security software. Moreover, completeness is vital for reasonable stop conditions in sequential diagnosis scenarios, e.g., if a single diagnosis remains after taking some measurements, only completeness implies that this diagnosis is the only possible minimal fault.\\ 
	\underline{\emph{Impact:}} Forgoing the requirement of completeness allows for a higher efficiency of diagnosis computation in many cases, as incomplete algorithms are often devised with a particular focus on performance, cf., e.g., \cite{feldman_computing_2008,abreu_low-cost_2009,li_computing_2002}.\\
	\underline{\emph{Relationship:}} If not carefully devised, incomplete algorithms will usually not be best-first.
	\item\label{enum:features:best-first} \textbf{Best-First Property:} \\
	\underline{\emph{Definition:}} 
	$A$ is \emph{generally best-first} iff it computes and outputs diagnoses in order according to a given sorting criterion (often: minimal cardinality or maximal probability); 
	$A$ is \emph{focused best-first} iff it is best-first only for a particular sorting criterion (often: minimal cardinality); 
	$A$ is \emph{only-best} iff it computes only the best diagnosis (if its type is single-solution, cf.\ Bullet~\ref{enum:features:type_of_output}) / diagnoses (if its type is multiple-solution, cf.\ Bullet~\ref{enum:features:type_of_output}) wrt.\ a particular property (often: minimum-cardinality);
	$A$ is \emph{best-subset-no-order} iff it computes a subset of all diagnoses including exactly the best diagnoses wrt.\ a particular property (often: minimal cardinality), but the diagnoses are not computed or output in best-first order;
	$A$ is \emph{any-first} iff it does not satisfy any of the above conditions and cannot guarantee any particular output order of diagnoses; 
	if $A$ is any-first, but uses heuristic techniques to approximate a particular order of the computed diagnoses, it is \emph{heuristic best-first}.\\
	\underline{\emph{Relevance:}} The best-first property is useful, e.g., if (one expects) there is a large number of diagnoses and the actual diagnosis is likely among the best diagnoses (e.g., when all components fail with an equal and small likelihood \cite{kleer_focusing_1991}), if focusing techniques are adopted where only the best subset of all diagnoses is further considered \cite{kleer_focusing_1991}, if informative samples for sequential diagnosis should be computed \cite{rodler_random_2022}, or if users intend to monitor the best diagnoses throughout the debugging process \cite{rodler_memory-limited_2022}.\\  
	\underline{\emph{Impact:}} Forgoing the best-first requirement usually leads to a higher efficiency of diagnosis computation, as any-first algorithms can use more efficient (e.g., depth-first \cite{shchekotykhin_sequential_2014} instead of breadth-first \cite{reiter_theory_1987} or uniform-cost \cite{rodler_interactive_2015}) diagnosis search techniques. Also, generally best-first methods might be significantly more expensive than related focused best-first ones (cf., e.g., \cite{rodler_memory-limited_2022}).\\
	\underline{\emph{Relationship:}} To the best of our knowledge, all generally best-first algorithms are conflict-dependent (cf.\ Bullet~\ref{enum:features:conflict-based}), i.e., rely on a systematic search based on conflicts (cf.\ Tab.~\ref{tab:algo_classification}). Moreover, compilation-based approaches (cf.\ Bullet~\ref{enum:features:conflict-based}) are usually only-best techniques wrt.\ minimum-cardinality diagnoses (cf.\ Tab.~\ref{tab:algo_classification}).
	\item \label{enum:features:type_of_output}\textbf{Type of Output:} \\
	\underline{\emph{Definition:}} $A$ is called \emph{multiple-solution} iff it can compute a set of two or more diagnoses per call; otherwise, if $A$ returns at most one diagnosis, it is called \emph{single-solution}.\\
	\underline{\emph{Relevance:}} The optimal algorithm choice wrt.\ this feature is trivial and depends on whether one or multiple diagnoses are required in a scenario. Note, most algorithms considered in Tab.~\ref{tab:algo_classification} can output multiple solutions. Clearly, any such algorithm can also be employed if only a single solution is desired. \\
	\underline{\emph{Impact:}} Single-solution techniques can be highly performant as they may use optimizations 
	that harm completeness   
	by manipulating the set of all solutions for the benefit of computational efficiency \cite{de_kleer_hitting_2011}. To allow for some degree of control over their performance, multiple-solution approaches are sometimes also configurable, e.g., to compute a number of exactly $k$ solutions, to stop after some timeout occurs, to prune a specified part of the search space, or to stop after a predefined number of search iterations has been performed \cite{rodler_memory-limited_2022,rodler_reuse_2020,rodler_statichs_2018,feldman_computing_2008,abreu_low-cost_2009}.\\
	\underline{\emph{Relationship:}} Single-solution methods are usually one-complete (cf.\ Bullet~\ref{enum:features:completeness}) and only-best (cf.\ Bullet~\ref{enum:features:best-first}), see Tab.~\ref{tab:algo_classification}. Simply put, when focusing on only one solution, approaches normally aim at finding the \emph{best} diagnosis wrt.\ some property among \emph{all} 
	minimal diagnoses. 
	\item \label{enum:features:conflict-based}\textbf{Conflict Dependency:} \\
	\underline{\emph{Definition:}} $A$ is \emph{conflict-dependent} iff it requires the computation of (minimal) conflicts;
	otherwise, i.e., if $A$ works without taking information about conflicts into account, it is \emph{direct}; a direct algorithm which translates 
	the DPI 
	into a target language and performs diagnosis computation based on this alternative problem representation is called \emph{compilation-based}.\\
	\underline{\emph{Relevance:}} If conflict computation or theorem proving is very expensive in a diagnostic scenario (cf., e.g., \cite{rodler_randomized_2021}), then direct algorithms are preferable, or even the only feasible approach. If a systematic exploration of diagnoses (allowing, e.g., inferences that all diagnoses with a specific property, say minimum-cardinality, are already computed) is desired \cite{kleer_focusing_1991} or the general best-first property (see bullet \ref{enum:features:best-first}) is relevant \cite{rodler_memory-limited_2022}, or the used method should be optimized for certain sequential diagnosis problems \cite{rodler_dynamichs_2020-1}, then a conflict-dependent approach might be the only viable choice. 
	When adopting a conflict-dependent method, it is important to note that an adequate conflict generation approach---which is sound and complete wrt.\ the computation of (minimal) conflicts as well as applicable to and performant for the DPI at hand---needs to be combined with the diagnosis algorithm. Choosing such an approach might not be an easy task for non-expert users.
	\\
	\underline{\emph{Impact:}} Compilation-based techniques often allow to answer important diagnostic queries (such as minimum-cardinality diagnosis computation) in polynomial time in the size of the compilation (which however might be of exponential size). Hence, these methods might be the best choice given that the DPI at hand is amenable to a compact compiled representation. Direct techniques, some of which (e.g., \cite{felfernig_efficient_2011,shchekotykhin_sequential_2014}) are based on the Duality Property (cf.\ Sec.~\ref{sec:basics}), sometimes allow to escape computational bottlenecks concerning memory consumption \cite{shchekotykhin_sequential_2014} or time \cite{feldman_computing_2008} by forgoing a \emph{systematic} enumeration of the diagnoses. Most of the algorithms in the literature appear to be conflict-dependent (cf.\ Tab.~\ref{tab:algo_classification}), and most (but not all, e.g., \cite{williams_conflict-directed_2007}) of them are based on the Hitting-set Property (cf.\ Sec.~\ref{sec:basics}); hence, there is a great selection of such methods, which cover numerous different combinations of other features, so that there will be a reasonable choice among them for many diagnosis tasks and applications.
	\\  
	\underline{\emph{Relationship:}} Considering the literature, it appears that conflict-dependency implies (if judgeable) the general applicability of an algorithm (cf.\ Bullet~\ref{enum:features:general_applicability} and Tab.~\ref{tab:algo_classification}). The reason for this is that, for diagnosis computation, the set of minimal conflicts is representative of a DPI (cf.\ \cite[Theorem~1]{de_kleer_mininimum_2009}), and thus can be seen as a kind of general abstraction from the DPI, which can be applied to any DPI. Hence, algorithms relying on this abstraction (usually) do not make assumptions about system specifics.
	%
	\item \label{enum:features:way_of_conflict_computation}\textbf{Way of Conflict Computation:} \\
	\underline{\emph{Definition:}} \emph{(Prerequisite: $A$ is conflict-dependent, cf.\ Bullet~\ref{enum:features:conflict-based})} $A$ is \emph{preliminary} iff it requires (a non-empty, non-singleton subset of) all minimal conflicts to be given as an input, or if it computes (a non-empty, non-singleton subset of) all minimal conflicts preliminarily, before the diagnosis computation starts; otherwise, if conflicts are computed on-demand in the course of diagnosis computation, $A$ is \emph{on-the-fly}.\\
	\underline{\emph{Relevance:}} If the prior generation of all minimal conflicts is feasible or even practical in a diagnosis scenario, there are highly efficient preliminary techniques available for diagnosis computation (cf., e.g., \cite{shi_exact_2010,de_kleer_hitting_2011}). These preliminary techniques can also benefit from insights of a significant body of research regarding the minimal hitting set problem (cf.\ \cite{gainer-dewar_minimal_2017} for an overview).  
	On-the-fly algorithms, on the other hand, are often still efficiently applicable even if preliminary conflict generation is infeasible. That said, it might in certain diagnostic use cases not be necessary to explicitly derive all (minimal) conflicts, e.g., in sequential diagnosis scenarios \cite{kleer_focusing_1991,rodler_active_2017} where only a subset of diagnoses is required per iteration. Some preliminary techniques (e.g., \cite{de_kleer_diagnosing_1987}) can be modified to act on-the-fly, but not all of them (e.g., ones that exploit the structure in the collection of minimal conflicts \cite{zhao_deriving_2015}). Any on-the-fly algorithm can be modified to be preliminary in a straightforward way (by pre-computing the collection of minimal conflicts and by choosing appropriate conflicts from this collection on-the-fly). 
	\\
	\underline{\emph{Impact:}} Forgoing the preliminary computation of the (full) set of minimal conflicts and intermixing conflict generation with diagnosis computation can allow to escape a combinatorial explosion and thus enhance the performance of diagnosis methods \cite{kleer_focusing_1991}. \\
	\underline{\emph{Relationship:}} Usually, preliminary algorithms do not incorporate mechanisms for generating minimal conflicts, but assume them to be given from the outset (e.g., \cite{lin_computation_2003,zhao_deriving_2015,de_kleer_hitting_2011}). For such methods, we cannot assess the features general applicability, black-box reasoning, and logics-agnosticism (cf.\ Bullets~\ref{enum:features:general_applicability}, \ref{enum:features:black-box_reasoning} and \ref{enum:features:logics-agnosticism}) because these methods do not directly use the DPI, but require some ``external'' technique to provide the required collection of conflicts, where the three said features above depend on the adopted conflict generation technique.
	\item \label{enum:features:focus_on_sequential_diagnosis}\textbf{Focus on Sequential Diagnosis:} \\
	\underline{\emph{Definition:}} 
	$A$ is \emph{sequential} iff it 
	provides mechanisms to address the sequential diagnosis problem (cf.\ Problem~\ref{prob:seq_diagnosis}), e.g., in terms of measurement proposal techniques or system knowledge update procedures after measurement actions; otherwise, 
	$A$ is \emph{one-shot}.\\
	\underline{\emph{Relevance:}} To solve a sequential diagnosis problem, algorithms devised specifically for this purpose will often be more practical than iteratively re-invoking a one-shot algorithm for the various DPIs (successively extended by new measurements, cf.\ Sec.~\ref{sec:basics}) during a sequential diagnosis session (cf., e.g., \cite{rodler_reuse_2020,siddiqi_sequential_2011}). Apart from that, the former techniques will often be directly applicable to a sequential diagnosis task, whereas a user might need to adapt the implementation of a one-shot algorithm to make it ready for sequential diagnosis. On the other hand, if sequential diagnosis is not the task in a diagnostic scenario, then a user is generally better off (wrt.\ efficiency, implementation complexity, etc.) when using one of the often less sophisticated (cf., e.g., \cite{rodler_dynamichs_2020-1}) one-shot techniques.
	\\
	\underline{\emph{Impact:}} Relying on sequential techniques will usually boost the performance of diagnosis computation in a sequential setting, but will generally also tend to worsen the performance in non-sequential settings.
	\\
	\underline{\emph{Relationship:}} Sequential techniques are usually sound, complete and stateful (cf.\ Bullets \ref{enum:features:soundness}, \ref{enum:features:completeness}, and \ref{enum:features:maintenance_of_state}, and Tab.~\ref{tab:algo_classification}) where the former two properties can be useful for diagnostic decision-making (e.g., measurement proposal, stop criteria) and the latter 
	can improve the time performance of an algorithm (cf.\ \cite{rodler_dynamichs_2020-1,de_kleer_diagnosing_1987}).
	\item \label{enum:features:maintenance_of_state} \textbf{Maintenance of State:} \\
	\underline{\emph{Definition:}} 
	$A$ is \emph{stateful} iff it can maintain its state when used throughout a sequential diagnosis process (e.g., by storing or reusing data structures, intermediate values, etc.); otherwise, $A$ is \emph{stateless}.\\
	\underline{\emph{Relevance:}} Since this feature 
	describes the internal workings of an algorithm, users might basically be indifferent whether the used diagnosis method is stateful or stateless. However, requirements wrt.\ the algorithm's performance may (ceteris paribus) have a bearing on the proper choice between stateful and stateless algorithms (see below). 
	\\
	\underline{\emph{Impact:}} When memory is the more critical resource, e.g., on small or mobile devices, stateless algorithms may be a way to trade more time for less space, whereas, when time is the more critical resource, stateful algorithms may be preferable \cite{rodler_dynamichs_2020-1,rodler_statichs_2018}. \\
	\underline{\emph{Relationship:}} Stateless algorithms are usually (but not always, cf.\ \cite{shchekotykhin_sequential_2014}) one-shot (cf.\ Bullet~\ref{enum:features:focus_on_sequential_diagnosis}), and algorithms that can be used in a stateful way are normally (but not always, cf.\ \cite{metodi_novel_2014}) sequential (cf.\ Bullet~\ref{enum:features:focus_on_sequential_diagnosis}). See Tab.~\ref{tab:algo_classification}.
	\item \label{enum:features:general_applicability}\textbf{General Applicability:} \\
	\underline{\emph{Definition:}} $A$ is \emph{generally applicable} iff it can be used for any diagnosis problem expressible by means of Reiter's theory \cite{reiter_theory_1987}, i.e., for any DPI as specified in Sec.~\ref{sec:basics}; otherwise, e.g., if $A$ makes certain assumptions about (e.g., the structure or some properties of) the tackled DPI, it is \emph{problem-dependent}.\\
	\underline{\emph{Relevance:}} The appropriate choice of diagnosis algorithm depends on its application area and scope. E.g., if only certain system types (such as circuits) are addressed, then a problem-dependent algorithm that considers and leverages the peculiarities of this system type will be the proper and often much more performant approach (cf.\ \cite{feldman_two-step_2006,siddiqi_sequential_2011,metodi_novel_2014}). If, on the other hand, a diagnosis system's intended use is for frequently changing application domains (e.g., in the Semantic Web context, where a multitude of different domains are modeled in terms of ontologies with highly heterogeneous content, structure, expressiveness, reasoning complexity and used logical languages \cite{shchekotykhin_interactive_2012,rodler_interactive_2015,rodler_memory-limited_2022}), problem-dependent techniques might not be eligible and generally applicable ones allow to deal with various diagnosis problems without modifying the diagnosis system.\\
	\underline{\emph{Impact:}} If general applicability is required, the price to pay for this is the use of general-purpose diagnostic techniques which naturally cannot match up in terms of performance to approaches geared to optimizing diagnostic efficiency for specific problem cases.\\
	\underline{\emph{Relationship:}} General applicability implies logics-agnosticism (cf.\ Bullet~\ref{enum:features:logics-agnosticism}) since an algorithm incapable of dealing with some (monotonic) logical language is, by definition, not generally applicable. However, there might be logics-agnostic techniques which exploit structural properties of a particular system type (regardless of how it is modeled), which are thus not generally applicable. Moreover, most (but not all, cf.\ \cite{de_kleer_diagnosing_1987}) of the generally applicable methods are black-box (wrt.\ reasoning), i.e., can use an arbitrary (sound and complete) inference mechanism (cf.\ Bullet~\ref{enum:features:black-box_reasoning}).  
	\item \label{enum:features:black-box_reasoning}\textbf{Black-Box Reasoning:} \\
	\underline{\emph{Definition:}} $A$ is \emph{black-box} 
	(wrt.\ reasoning) iff it uses a reasoner as a black-box oracle 
	(for consistency or model checking)
	and can use an arbitrary (sound and complete) reasoner for the logical language used to express the DPI; otherwise, if $A$ requires additional computations or mechanisms (e.g., operations pertinent to a specific problem representation \cite{torasso_model-based_2006,darwiche_decomposable_2001} or bookkeeping techniques \cite{de_kleer_diagnosing_1987}) 
	from a reasoner beyond the main reasoning result, it is \emph{reasoner-dependent}.\footnote{Methods we call reasoner-dependent are sometimes also referred to as \emph{glass-box} techniques \cite{parsia_debugging_2005,horridge_justification_2011,kalyanpur_debugging_2006}.}\\
	\underline{\emph{Relevance:}} If the logic used to model the diagnosed system is stable in an application area, then reasoner-dependent approaches might be the better choice as they might be advantageous in terms of diagnostic efficiency 
	(given a suitable ``glass-box'' reasoner for the respective logic).
	E.g., when DPIs expressible by means of propositional logic are the target use case of a diagnosis system, then a reasoner-dependent algorithm based on propositional logic might be preferable to a black-box one with otherwise equal features. If, on the other hand, different formalisms might be used to describe the faulty system \cite{rodler_memory-limited_2022}, black-box techniques can be more expedient. They can always simply use the best reasoner for the particular problem at hand without needing to incorporate any modifications into the reasoner (or the diagnosis algorithm)---unlike reasoner-dependent approaches, which require the incorporation of the necessary additional mechanisms into any adopted reasoner. And, performances of various reasoners might differ substantially \cite{goncalves_owl_2013}. As a rule of thumb, if the performance for one \emph{fixed} system description language should be maximized, then reasoner-dependent approaches tend to be more favorable, whereas black-box methods tend to be preferable if the performance over \emph{variable} modeling languages should be optimized.   
	\\
	\underline{\emph{Impact:}} Reasoner-dependent techniques can lead to an improved time performance \cite{horridge_justification_2011,kalyanpur_debugging_2006}, e.g., when reasoners extract conflicts as a byproduct of consistency checks \cite{horridge_justification_2011,kalyanpur_debugging_2006} or store supporting environments\footnote{Roughly: sets of logical sentences sufficient for the entailment to hold. Environments are also termed \emph{justifications} \cite{horridge_justification_2011}.} for derived entailments \cite{de_kleer_diagnosing_1987}, but might also incur memory overheads \cite{kalyanpur_debugging_2005}. Advantages of black-box methods are, e.g., their robustness (no sophisticated, and potentially error-prone, modifications of complex reasoning algorithms), their simplicity (internals of reasoner irrelevant), their flexibility (e.g., black-box methods can use a portfolio reasoning approach by switching to the most efficient reasoner in a simple plug-in fashion depending on the language used to describe the diagnosed system \cite{romero_more_2012}), and their up-to-dateness (black-box methods can directly benefit from advances in the general research on automated reasoning).\\ 
	\underline{\emph{Relationship:}} There are no general implications on other features resulting from the presence or absence of the black-box property; however, it is often (but not always) the case that black-box techniques are also generally applicable, logics-agnostic, sound, complete and multiple-solution (cf.\ Bullets~\ref{enum:features:soundness}, \ref{enum:features:completeness}, \ref{enum:features:general_applicability}, \ref{enum:features:logics-agnosticism} and \ref{enum:features:type_of_output}).
	\item \label{enum:features:logics-agnosticism}\textbf{Logics-Agnosticism:} \\
	\underline{\emph{Definition:}} $A$ is \emph{logics-agnostic} iff it can deal with DPIs expressed by arbitrary (monotonic) logics; otherwise, $A$ is \emph{logics-dependent}.\\
	\underline{\emph{Relevance:}} If a diagnosis approach is intended to be used with only one \emph{fixed} system description language, then a user should choose the method with (expected/reported) best performance for the faced diagnostic task, regardless of whether it is logics-agnostic or -dependent. Logics-dependent approaches, however, can offer very attractive features, e.g., compilation-based approaches (e.g., \cite{metodi_novel_2014,torasso_model-based_2006,darwiche_decomposable_2001}) can often compute diagnoses in polynomial time once the DPI has been compiled into a target representation; however, they are restricted to propositional logic system descriptions. If a diagnosis approach needs to deal with \emph{diverse} system modeling languages, the adoption of a  logics-agnostic method might be the only choice. \\
	\underline{\emph{Impact:}} As our literature study suggests, logics-dependent approaches are usually particularly attractive in terms of performance. The reason for this is that these methods often use sophisticated (e.g., representation, optimization or reasoning) techniques specific to one logic, which is propositional (Horn) logic for all logics-dependent approaches considered in Tab.~\ref{tab:algo_classification}. \\
	\underline{\emph{Relationship:}} The property of logics-agnosticism appears to come along with the black-box property (cf.\ Bullet~\ref{enum:features:black-box_reasoning}) in most (but not all, cf.\ \cite{de_kleer_diagnosing_1987}) cases. Furthermore, logics-agnostic techniques are often generally applicable (cf.\ Bullet~\ref{enum:features:general_applicability}). 
	\item \label{enum:features:space_efficiency}\textbf{Space Efficiency:} \\
	\underline{\emph{Definition:}} $A$ is \emph{space-efficient} iff its worst-case memory complexity is polynomial in the input size; otherwise, $A$ is \emph{space-inefficient}.\\
	\underline{\emph{Relevance:}} 
	If time is the resource to be minimized and the diagnostic task is expected to manage with the available memory, then space-inefficient methods can be the better choice, due to their generally lower time complexity.
	If, on the other hand, the available memory capacity of a device is low (such as with IoT or smaller mobile devices) or the diagnostic task is expected to be memory-intensive (e.g., if diagnosis cardinality is likely to be high \cite{shchekotykhin_sequential_2014}), then space-efficient algorithms might be the only viable approach (as memory consumption increases exponentially with the size of diagnoses for many space-inefficient techniques). Note, only few space-efficient diagnosis computation strategies have been proposed in literature, thus there is not plenty of choice for the user, which is why a trade-off between space-efficiency and other properties might be necessary. Only recently, a space-efficient algorithm exhibiting all desirable properties wrt.\ the features discussed here, along with a reasonable time performance, has been suggested \cite{rodler_memory-limited_2022}.  
	\\
	\underline{\emph{Impact:}} Space-efficiency is often bought for a higher (empirical or theoretical) time complexity \cite{rodler_memory-limited_2022}, or for a dropping of other desirable properties such as best-firstness \cite{shchekotykhin_sequential_2014,felfernig_efficient_2011} or completeness \cite{abreu_low-cost_2009}.\\ 
	\underline{\emph{Relationship:}} 
	With the exception of the algorithm proposed in \cite{rodler_memory-limited_2022}, 
	space-efficiency appears to be achievable only for algorithms that do not guarantee a best-first diagnosis computation (cf.\ Bullet \ref{enum:features:best-first}), or that require a preliminary computation of conflicts (cf.\ Bullet~\ref{enum:features:way_of_conflict_computation}) and whose complexity thus does not take into account the conflict generation phase. 
\end{enumerate}
\noindent\textbf{Remarks:}
\begin{itemize}[noitemsep,topsep=0pt]
	\item We do not propose a feature concerning the time complexity. This is due to well-known complexity results \cite{eiter_complexity_1995,bylander_computational_1991}, which imply (unless P = NP) that there cannot be an algorithm that computes at least two minimal diagnoses and generally finishes in polynomial time; this holds even if reasoning (consistency checking) is in P, which however is already NP-complete if (only) propositional logic models are used (let alone more expressive logics such as Description Logics \cite{baader_description_2007}). 
	\item The list of proposed features is certainly not an exhaustive account of all possible properties diagnosis computation algorithms might have. There are further conceivable aspects from the
	\emph{(a)}~\emph{theoretical viewpoint}, such as whether an algorithm uses abstractions or alterations of a DPI, or whether computations are executed in a distributed or centralized way (cf.\ \cite[Sec.~4]{rodler_dynamichs_2020-1}), 
	\emph{(b)}~\emph{empirical viewpoint}, such as whether an algorithm was experimentally evaluated, which dataset from which domain was used to evaluate an algorithm, or which other methods an algorithm was compared against, 
	\emph{(c)}~\emph{presentation viewpoint}, such as whether formal proofs for algorithm properties are given, or from the 
	\emph{(d)}~\emph{pragmatic viewpoint}, such as whether there are freely accessible implementations of or tools relying on an algorithm. 
	Exploring other features like these is on our future work agenda.    
\end{itemize}

\section{Classification of Existing Works}
\label{sec:classification}
Tab.~\ref{tab:algo_classification} gives a classification of several existing works based on their characteristics wrt.\ the features suggested in Sec.~\ref{sec:features}. 
\noindent\textbf{Remarks:}
\begin{itemize}[noitemsep,topsep=0pt]
	\item The table can be read row-wise to inspect the features of the diagnostic techniques, and column-wise to find methods with certain characteristics wrt.\ the features.
	\item We assessed the algorithms enumerated in the table as they are described in the respective cited work, without assuming any modifications or extensions.
	\item The list of algorithms studied in the table raises no claim to completeness. Rather, the idea is to illustrate the use(fulness) of the presented features for algorithm assessment and comparison by presenting the properties of \emph{some} important methods in literature. We plan to 
	analyze further 
	ones as part of our future work. 
\end{itemize}
  
\section{Conclusions}
\label{sec:conclusion}
This work proposes a taxonomy for diagnosis computation algorithms, with the intention of helping researchers and practitioners in assessing, comparing, and selecting diagnostic techniques for their tasks and purposes. More specifically, we present a set of 12 crucial features of diagnosis techniques and classify some important existing methods based on these. In our study of the works in the literature, we observed that, for some algorithms, it was relatively hard to determine their properties regarding the proposed taxonomy since various algorithmic aspects are often left implicit or not addressed at all. Moreover, even if properties are discussed, not all works provide formal proofs of these (which is certainly not least because of strict space restrictions at many publication venues). Hence, we encourage authors (whenever possible) to \emph{explicitly} discuss the material properties of their proposed diagnostic approaches, for which we hope the suggested taxonomy will constitute a useful basis and guideline. Making algorithm characteristics explicit, clear and accessible will certainly help other researchers put novel works appropriately into the context of existing works, and sticking to shared assessment and categorization criteria while doing so can have several advantages. Examples are \emph{(a)}~fair empirical evaluations that contrast methods which are actually comparable (e.g., comparing an incomplete method against a complete one wrt.\ performance might under circumstances be pointless, as the methods simply accomplish different things and have different use cases), \emph{(b)}~the potential inspiration for and detection of open research questions 
(e.g., find an algorithm which has a particular subset of the features, which are not met by any of the existing algorithms), \emph{(c)}~an easier, faster and more informed finding of a suitable algorithm for a particular purpose (e.g., is there a space-efficient method for a mobile device that is at the same time sound, complete and best-first?), 
\emph{(d)}~the better understanding of the evolution and reality in research and practice (e.g., that certain feature combinations are the reason why some techniques are not used in some application domains while they are state-of-the-art in others), 
or \emph{(e)}~the realization that basically all algorithms have their right to exist, as they cover a wide variety of feature combinations and thus address a broad range of diagnostic problem scenarios, and that algorithms ``superseding'' others due to performance improvements most often achieve this at the cost of losing some desirable properties. 

\begin{sidewaystable*}[t]
	\scriptsize
	\centering
	\rowcolors{2}{white}{gray!20}
		\begin{tabular}{@{}>{\columncolor{white}[0pt][\tabcolsep]}lrr|llll|ll|ll|lll|>{\columncolor{white}[\tabcolsep][0pt]}l@{}}
			\hline
			\rowcolor{gray!50}
			\multicolumn{3}{@{}>{\columncolor{white}[0pt][\tabcolsep]}c|}{\textbf{Technique}}  &     \multicolumn{12}{c}{\textbf{Features}} 
			\\
			\hline
			\rowcolor{gray!50}
			\multicolumn{3}{@{}>{\columncolor{white}[0pt][\tabcolsep]}c|}{}  &     \multicolumn{4}{c|}{\textbf{Output Qualities}} & \multicolumn{2}{c|}{\textbf{Way of Computation}} & \multicolumn{2}{c|}{\textbf{Seq.\ Diag.\ Context}} & \multicolumn{3}{c|}{\textbf{Application Context}} & \multicolumn{1}{c}{\textbf{Performance}} 
			\\ 
			\rowcolor{gray!50}
			Name & Year & Work                   					& SOUND  			& COMPL      & BEST-F  & MULT-SOL     & CONF-DEP         & O-T-FLY & SEQ 	& STATE & GEN-APPL & BL-BOX-REAS & ANY-LOGIC                  & POLY-SPACE \\ \hline
			GDE & 1987  &\cite{de_kleer_diagnosing_1987}  					& $\checkmark$      & $\checkmark_{(\text{all})}$          & $\checkmark_{(\text{gen})}$          &    $\checkmark$   & $\checkmark$                & $\times$         & $\checkmark$   	& $\checkmark$     & $\checkmark$        & $\times_{(\text{bk})}$           & $\checkmark$                          & $\times$                  \\ 
			HS-Tree & 1987  &\cite{reiter_theory_1987}    					& $\checkmark$      & $\checkmark_{(\text{all})}$          & $\checkmark_{(\text{foc}=\text{mc})}$    & $\checkmark$       & $\checkmark$                & $\checkmark$         & $\times$   	& $\times$     & $\checkmark$        & $\checkmark$           & $\checkmark$                          & $\times$                 \\
			HS-DAG & 1989  &\cite{greiner_correction_1989}					& $\checkmark$      & $\checkmark_{(\text{all})}$          & $\checkmark_{(\text{gen})}$     & $\checkmark$      & $\checkmark$                & $\checkmark$         & $\times$   	& $\times$     & $\checkmark$        & $\checkmark$           & $\checkmark$                          & $\times$                 \\
			DIAGNOSE & 1994  &\cite{hou_theory_1994}      					& $\checkmark$      & $\checkmark_{(\text{all})}$          & $\times$      & $\checkmark$      & $\checkmark$                & $\times$         & $\checkmark$   	& $\checkmark$     & $\checkmark$        & $\checkmark$           & $\checkmark$                          & $\times$                \\
			HST & 2001  &\cite{wotawa_variant_2001}       					& $\checkmark$      & $\checkmark_{(\text{all})}$          & $\checkmark_{(\text{foc}=\text{mc})}$   & $\checkmark$  & $\checkmark$                & $\checkmark$         & $\times$   	& $\times$     & $\checkmark$        & $\checkmark$           & $\checkmark$                          & $\times$                 \\
			DNNF & 2001  &\cite{darwiche_decomposable_2001}					& $\checkmark$      &$\checkmark_{(\text{p}=\text{mc})}$   & $\checkmark_{(\text{only}=\text{mc})}$  & $\checkmark$   & $\times_{(\text{cp-b} = \text{DNNF})}$       & na        & $\times$   	& ?     & $\times$   & $\times$           & $\times_{(\text{PL})}$                     & $\times$                \\
			Genetic Alg. & 2002  &\cite{li_computing_2002}					& $\times$      & $\times$          & $\times$      & $\checkmark$     & $\checkmark$                & $\times$         & $\times$   	& $\times$     & na       & na          & na                         & ?                 \\
			BHS-Tree & 2003  &\cite{lin_computation_2003} 					& $\checkmark$      & $\checkmark_{(\text{all})}$          & $\times$      & $\checkmark$      & $\checkmark$                & $\times$         & $\times$   	& $\times$     & na       & na          & na                         & $\times$                \\
			Bool.\ Alg. & 2003  &\cite{lin_computation_2003}					& $\checkmark$      & $\checkmark_{(\text{all})}$          & $\times$     & $\checkmark$     & $\checkmark$                & $\times$         & $\times$   	& $\times$     & na       & na          & na                         & $\times$                  \\
			HSSE-Tree & 2006  &\cite{xiangfu_method_2006} 					& $\checkmark$      & $\checkmark_{(\text{all})}$          & $\checkmark_{(\text{foc}=\text{mc})}$  & $\checkmark$  & $\checkmark$                & $\times$         & $\times$   	& $\times$     & na       & na          & na                         & $\times$                  \\
			HA*  & 2006  &\cite{feldman_two-step_2006}    					& $\checkmark$      & $\checkmark_{(\text{one}=\text{mc})}$         & $\checkmark_{(\text{only}=\text{mc})}$  & $\times$ & $\times_{(\text{cp-b} = \text{h-DNF})}$      & na        & $\times$     	& ?     & $\times_{(\text{circ})}$   & $\checkmark$           & $\times_{(\text{PL})}$             		& $\times$                  \\
			OBDD  & 2006  &\cite{torasso_model-based_2006}					& $\sim^{\scaleobj{.75}{(1)}}$ 			& $\checkmark_{(\text{all})}$          & $\checkmark_{(\text{only}=\text{mc})}$  & $\checkmark$ & $\times_{(\text{cp-b} = \text{OBDD})}$       & na        & $\checkmark$   	& $\checkmark$     & $\times$  & $\times$          & $\times_{(\text{PL/HL})}$ 					& $\times$                  \\
			CDA*  & 2007  &\cite{williams_conflict-directed_2007}			& $\checkmark$      & $\times$          & $\checkmark_{(\text{gen})}$       & $\checkmark$    & $\checkmark$                & $\checkmark$         & $\times$  		& ?     & $\checkmark$        & $\checkmark$           & $\checkmark$                          & $\times$                 \\
			SAFARI  & 2008  &\cite{feldman_computing_2008}                   & $\sim^{\scaleobj{.75}{(2)}}$ & $\times$          & $\times$       & $\checkmark$     & $\times_{(\text{dir})}$                & na        & $\times$   	& $\times$     & $\checkmark$        & $\checkmark$           & $\checkmark$                          & $\times$                \\
			STACCATO  & 2009  &\cite{abreu_low-cost_2009}                   	& ?      & $\sim^{\scaleobj{.75}{(3)}}$ & $\times_{(\text{heur})}$      & $\checkmark$    & $\checkmark$                & $\times$         & $\times$   	& $\times$     & na       & na          & na                         & $\checkmark$                \\
			NGDE  & 2009  &\cite{de_kleer_mininimum_2009}                    & $\checkmark$      & $\checkmark_{(\text{p}=\text{mc})}$   & $\checkmark_{(\text{only}=\text{mc})}$ & $\checkmark$   & $\checkmark$                & $\checkmark$         & $\times$   	& ?     & $\checkmark$        & $\times_{(\text{bk})}$      & $\checkmark$                          & ?$^{\scaleobj{.75}{(4)}}$            \\
			Recurs.\ MHS  & 2010  &\cite{shi_exact_2010}         			& $\checkmark$      & $\checkmark_{(\text{one}=\text{mc})}$    & $\checkmark_{(\text{only}=\text{mc})}$  & $\times$  & $\checkmark$                & $\times$         & $\times$   	& na    & na       & na          & na                         & $\sim^{\scaleobj{.75}{(5)}}$                \\
			SDA  & 2011  &\cite{siddiqi_sequential_2011}                     & $\checkmark$      & $\checkmark_{(\text{one}=\text{SD-sol})}$         & $\checkmark_{(\text{only}=\text{SD-sol})}$   & $\times$    & $\times_{(\text{cp-b} = \{\text{BN, d-DNNF}\})}$ & na        & $\checkmark$   	& $\checkmark$     & $\times_{(\text{circ})}$   & $\times$           & $\times_{(\text{PL})}$             		& $\times^{\scaleobj{.75}{(6)}}$          \\
			cminc  & 2011  &\cite{de_kleer_hitting_2011}                     & $\checkmark$      & $\checkmark_{(\text{one}=\text{mc})}$    & $\checkmark_{(\text{only}=\text{mc})}$  & $\times$  & $\checkmark$                & $\times$         & $\times$   	& na    & na       & na          & na                         & ?$^{\scaleobj{.75}{(7)}}$             \\
			FastDiag  & 2011  &\cite{felfernig_efficient_2011}               & $\checkmark$      & $\checkmark_{(\text{all})}$          & $\times$      & $\checkmark$     & $\times_{(\text{dir})}$                & na        & $\times$   	& $\times$     & $\checkmark$        & $\checkmark$           & $\checkmark$                          & $\checkmark$                 \\
			SDE & 2012  &\cite{stern_exploring_2012}                        	& $\checkmark$      & $\checkmark_{(\text{all})}$          & $\checkmark_{(\text{foc}=\text{mc})}$  & $\checkmark$   & $\checkmark$                & $\checkmark$         & $\times$   	& $\times$     & $\checkmark$        & $\checkmark$           & $\checkmark$                          & $\times$                 \\
			Improved Bool.\ Alg.$^{\scaleobj{.75}{(*)}}$ & 2012  &\cite{pill_optimizations_2012}  	& $\checkmark$      & $\checkmark_{(\text{all})}$          & $\checkmark_{(\text{bsno})}$  & $\checkmark$   & $\checkmark$                & $\times$         & $\times$   	& na    & na       & na          & na                         & $\times$                 \\
			Inv-HS-Tree  & 2014  &\cite{shchekotykhin_sequential_2014}       & $\checkmark$      & $\checkmark_{(\text{all})}$          & $\times$     & $\checkmark$      & $\times_{(\text{dir})}$                & na        & $\checkmark$   	& $\times$     & $\checkmark$        & $\checkmark$           & $\checkmark$                          & $\checkmark$                 \\
			SATbD  & 2014  &\cite{metodi_novel_2014}                     	& $\checkmark$      & $\checkmark_{(\text{p}=\text{mc})}$   & $\checkmark_{(\text{only}=\text{mc})}$  & $\checkmark$   & $\times_{(\text{cp-b} = \text{SAT})}$        & na        & $\times$   	& $\checkmark$     & $\times_{(\text{circ})}$   & $\checkmark$           & $\times_{(\text{PL})}$             		& ?                 \\
			Increm-distrib-MHS$^{\scaleobj{.75}{(*)}}$  & 2015  &\cite{zhao_deriving_2015}             	& $\checkmark$      & $\checkmark_{(\text{all})}$          & $\times$     & $\checkmark$       & $\checkmark$                & $\times$         & $\checkmark$   	& $\checkmark$     & na       & na          & na                         & $\times$                \\
			Unif-cost HS-Tree$^{\scaleobj{.75}{(*)}}$  & 2015  &\cite{rodler_interactive_2015}       & $\checkmark$      & $\checkmark_{(\text{all})}$          & $\checkmark_{(\text{gen})}$     & $\checkmark$       & $\checkmark$                & $\checkmark$         & $\times$   	& $\times$     & $\checkmark$        & $\checkmark$           & $\checkmark$                          & $\times$                \\
			Parallel HS-Tree  & 2016  &\cite{jannach_parallel_2016}          & $\checkmark$      & $\checkmark_{(\text{all})}$          & $\checkmark_{(\text{foc}=\text{mc})}$  & $\checkmark$   & $\checkmark$                & $\checkmark$         & $\times$   	& $\times$     & $\checkmark$        & $\checkmark$           & $\checkmark$                          & $\times$                 \\
			StaticHS  & 2018  &\cite{rodler_statichs_2018}                  	& $\checkmark$      & $\checkmark_{(\text{all})}$          & $\checkmark_{(\text{gen})}$    & $\checkmark$       & $\checkmark$                & $\checkmark$         & $\checkmark$   	& $\checkmark$     & $\checkmark$        & $\checkmark$           & $\checkmark$                          & $\times$                 \\
			DynamicHS  & 2020  &\cite{rodler_reuse_2020}                		& $\checkmark$      & $\checkmark_{(\text{all})}$          & $\checkmark_{(\text{gen})}$     & $\checkmark$       & $\checkmark$                & $\checkmark$         & $\checkmark$   	& $\checkmark$     & $\checkmark$        & $\checkmark$           & $\checkmark$                          & $\times$                \\
			RBF-HS  & 2022  &\cite{rodler_memory-limited_2022}               & $\checkmark$      & $\checkmark_{(\text{all})}$          & $\checkmark_{(\text{gen})}$     & $\checkmark$     & $\checkmark$                & $\checkmark$         & $\times$   	& $\times$     & $\checkmark$        & $\checkmark$           & $\checkmark$                          & $\checkmark$                  \\
			HBF-HS  & 2022  &\cite{rodler_memory-limited_2022}               & $\checkmark$      & $\checkmark_{(\text{all})}$          & $\checkmark_{(\text{gen})}$       & $\checkmark$     & $\checkmark$                & $\checkmark$         & $\times$   	& $\times$     & $\checkmark$        & $\checkmark$           & $\checkmark$                          & $\times$          \\
			Heuristic Inv-HS-Tree$^{\scaleobj{.75}{(*)}}$  & 2022  &\cite{rodler_random_2022}             & $\checkmark$      & $\checkmark_{(\text{all})}$          & $\times_{(\text{heur})}$       & $\checkmark$    & $\times_{(\text{dir})}$                & na        & $\times$   	& $\times$     & $\checkmark$        & $\checkmark$           & $\checkmark$                          & $\checkmark$            \\ \hline  
		\end{tabular}
		\caption{\footnotesize Classification of some existing diagnosis computation algorithms based on their characteristics wrt.\ the features proposed in Sec.~\ref{sec:features}. Table rows are sorted by the year of publication of the algorithms. Features (table columns) are thematically grouped as described in Sec.~\ref{sec:features}. 
			\emph{(Column meanings:)} 
			SOUND...is the algorithm sound? (Bullet~\ref{enum:features:soundness}); 
			COMPL...is the algorithm complete? (Bullet~\ref{enum:features:completeness}); 
			BEST-F...is the algorithm best-first? (Bullet~\ref{enum:features:best-first}); 
			MULT-SOL...is the algorithm multiple-solution? (Bullet~\ref{enum:features:type_of_output});
			CONF-DEP...is the algorithm conflict-dependent? (Bullet~\ref{enum:features:conflict-based}); O-T-FLY...does the algorithm (if applicable) compute conflicts on-the-fly? (Bullet~\ref{enum:features:way_of_conflict_computation}); SEQ...is the algorithm sequential? (Bullet~\ref{enum:features:focus_on_sequential_diagnosis}); 
			STATE...is the algorithm stateful? (Bullet~\ref{enum:features:maintenance_of_state}); GEN-APPL...is the algorithm generally applicable? (Bullet~\ref{enum:features:general_applicability}); BL-BOX-REAS...is the algorithm black-box wrt.\ reasoning? (Bullet~\ref{enum:features:black-box_reasoning}); ANY-LOGIC...is the algorithm logics-agnostic? (Bullet~\ref{enum:features:logics-agnosticism}); POLY-SPACE...is the algorithm space-efficient? (Bullet~\ref{enum:features:space_efficiency}). 
			\emph{(Symbol meanings:)} $\checkmark$...yes; $\sim$...under certain circumstances; $\times$...no; $\checkmark_{(\text{all})}$...all-complete; $\checkmark_{(\text{p} = X)}$...property-complete (wrt.\ property $X$); $\checkmark_{(\text{one} = X)}$...one-complete (wrt.\ property $X$); $\checkmark_{(\text{gen})}$...generally best-first; $\checkmark_{(\text{foc} = X)}$...focused best-first (wrt.\ property $X$); $\checkmark_{(\text{only} = X)}$...only-best (wrt.\ property $X$); $\checkmark_{(\text{bsno} = X)}$...best-subset-no-order (wrt.\ property $X$); $\times_{(\text{heur})}$...heuristic best-first; $\times_{(\text{cp-b} = X)}$...compilation-based (using target language $X$); $\times_{(\text{dir})}$...direct; na...not applicable; $\times_{(\text{circ})}$...specific to circuit diagnosis problems; $\times_{(\text{bk})}$...uses bookkeeping mechanism for reasoning (ATMS for \protect\cite{de_kleer_diagnosing_1987}, HTMS for \protect\cite{de_kleer_hitting_2011}); $\times_{(\text{PL})}$...specific to propositional logic; $\times_{(\text{PL/HL})}$...specific to propositional logic / Horn logic; ?...unknown. 
			\emph{(Table notes:)}
			$^{(*)}$...algorithm name given in the table not used in the original work; 
			$^{(1)}$...sound wrt.\ all diagnoses, but not wrt.\ all \emph{minimal} diagnoses;
			$^{(2)}$...unsound in most efficient configuration; can be parameterized to be sound; 
			$^{(3)}$...incomplete in most efficient configuration; can be parameterized to be complete; 
			$^{(4)}$...the diagnosis search is depth-first, i.e., requires linear space, but unclear if the used HTMS is polynomial space; 
			$^{(5)}$...polynomial-space minimal hitting set computation, but assumes a given (i.e., preliminarily computed) set of conflicts; 
			$^{(6)}$...d-DNNF is less succinct (i.e., larger in general) than DNNF \protect\cite[Sec.~2]{darwiche_new_2004}, and compilation to DNNF may result in exponential-size compilations \protect\cite{darwiche_decomposable_2001}; 
			$^{(7)}$...uses depth-first search, which requires linear space, but unclear if FullReduce() function \protect\cite{de_kleer_hitting_2011}, which has to consider all conflicts at once, is polynomial space.}
		\label{tab:algo_classification}
	\end{sidewaystable*}

\fontsize { 8.5pt }{ 8.5pt } 
\selectfont

\end{document}